 \documentclass[pmlr,twocolumn,10pt]{jmlr} 

\usepackage{rotating}
\usepackage{longtable}
\usepackage{natbib}

\usepackage{booktabs}
\usepackage{siunitx}

\newcommand{\model}{UniQA\xspace}
\newcommand\boldred[1]{\textcolor{red}{\textbf{#1}}}

\newcommand{\hide}[1]{}

\usepackage{enumitem}
\usepackage{float}
\usepackage{lipsum}

\newcommand{\yb}{\mathbf{y}}

\theorembodyfont{\upshape}
\theoremheaderfont{\scshape}
\theorempostheader{:}
\theoremsep{\newline}

\jmlrvolume{LEAVE UNSET}
\jmlryear{2021}
\jmlrsubmitted{LEAVE UNSET}
\jmlrpublished{LEAVE UNSET}
\jmlrworkshop{Machine Learning for Health (ML4H) 2021} 

\title[Question Answering for Complex EHR Database using Unified Encoder-Decoder Architecture]{Question Answering for Complex Electronic Health Records Database using Unified Encoder-Decoder Architecture}

\author{%
\Name{Seongsu Bae}
\Email{seongsu@kaist.ac.kr}\\
\addr KAIST AI / Daejeon, South Korea
\AND
\Name{Daeyoung Kim}
\Email{daeyoung.k@kaist.ac.kr}\\
\addr KAIST AI / Daejeon, South Korea
\AND
\Name{Jiho Kim}
\Email{jiho283@kaist.ac.kr}\\
\addr KAIST AI / Daejeon, South Korea
\AND
\Name{Edward Choi} \Email{edwardchoi@kaist.ac.kr}\\
\addr KAIST AI / Daejeon, South Korea
}

\begin{document}

\maketitle

\vspace{-5mm}
\begin{abstract}

An intelligent machine that can answer human questions based on electronic health records (EHR-QA) has a great practical value, such as supporting clinical decisions, managing hospital administration, and medical chatbots.
Previous table-based QA studies focusing on translating natural questions into table queries (NLQ2SQL), however, suffer from the unique nature of EHR data due to complex and specialized medical terminology, hence increased decoding difficulty.
In this paper, we design \model, a unified encoder-decoder architecture for EHR-QA where natural language questions are converted to queries such as SQL or SPARQL.
We also propose input masking (IM), a simple and effective method to cope with complex medical terms and various typos and better learn the SQL/SPARQL syntax.
Combining the unified architecture with an effective auxiliary training objective, \model demonstrated a significant performance improvement against the previous state-of-the-art model for MIMICSQL* (14.2\% gain), the most complex NLQ2SQL dataset in the EHR domain, and its typo-ridden versions ($\approx$ 28.8\%  gain).
In addition, we confirmed consistent results for the graph-based EHR-QA dataset, MIMICSPARQL*.
\end{abstract}

\section{Introduction}
\label{sec:intro}

\begin{figure}[t]
\floatconts
    {fig:scenarios}
    {\caption{An example of natural language question (NLQ) and SQL query pairs for EHR question answering.}\vspace{-1em}}
    {\vspace{1em}\includegraphics[width=1.0\columnwidth]{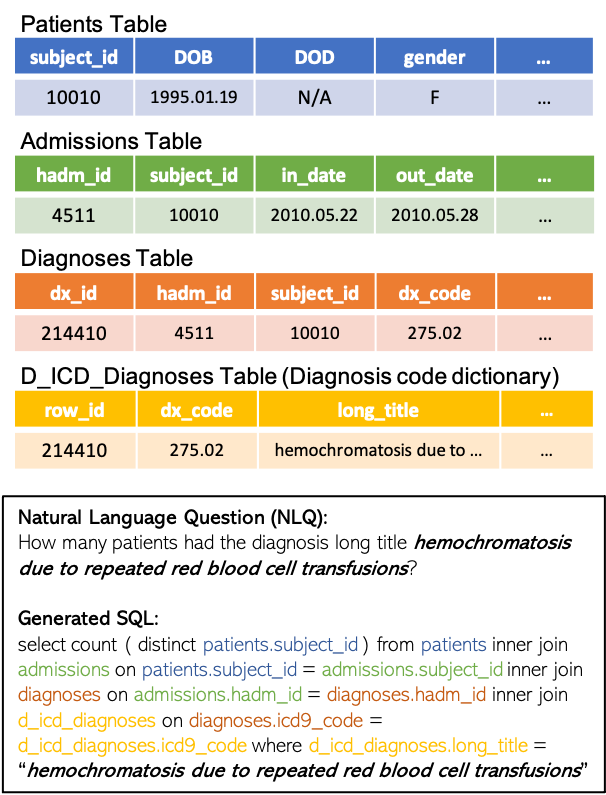}\vspace{-1em}}
\end{figure}

Electronic health records (EHR) consist of real-world clinical data (e.g., patient diagnoses, medications, lab results), usually stored in a complex relational database (RDB) such as MIMIC-III \citep{johnson2016mimic} and eICU \citep{pollard2018eicu}.
For healthcare providers or ordinary users, performing information retrieval or knowledge inference from such massive hospital database systems is not a trivial task.
They not only must learn to use an appropriate query language (e.g., SQL, SPARQL) but also learn the schema and the corresponding values in the hospital RDB. 
Therefore developing an intelligent EHR Question Answering (EHR-QA) model that can answer human-level questions from the database has immense practical values such as supporting clinical decisions, managing hospital administration, and medical chatbots.

Recently, translating natural language questions into corresponding SQL queries (\textit{i.e.} NLQ2SQL) \citep{zhong2017seq2sql, hwang2019comprehensive, wang2020rat} has become the dominant approach in QA over relational databases.
In the healthcare domain, \cite{wang2020text} was the first attempt to construct MIMICSQL, a large-scale NLQ2SQL dataset built from the open-source EHR dataset MIMIC-III~\citep{johnson2016mimic}, which was followed by MIMICSQL* and MIMICSPARQL*~\citep{park2021knowledge}, more refined EHR-QA datasets derived from MIMICSQL.

Unlike the general domain QA, EHR-QA faces a unique challenge due to the distinguished nature of EHR data.
Figure~\ref{fig:scenarios} shows an illustrative example of EHR-QA, which presents a pair of the natural language question and the corresponding SQL query and multiple relational tables required to retrieve it.
Although the input question (\textit{i.e.} ``\textit{How many patients had the diagnosis...?}'') is of natural syntax, it contains long and domain-specific terms, such as ``\textit{hemochromatosis due to repeated red blood cell transfusions}'', which can lead to increased decoding difficulty and various typos, such as missing or reversed letters.

In this work, we propose \model, an effective EHR-QA model based on the unified encoder-decoder architecture to cope with the discussed practical and realistic challenge.
Using a combination of the unified \textit{Encoder-as-Decoder} architecture, input token masking, and the value recovering technique, \model was able to achieve the state-of-the-art EHR-QA performance on MIMICSQL* as well as robustness against its variants with various typos for input questions.
Furthermore, \model showed consistent empirical results for the graph-based EHR-QA dataset, MIMICSPARQL*.

The contributions of this work can be summarized as follows:
\begin{itemize}
    \item We propose \model, a unified \textit{Encoder-as-Decoder} model to answer EHR-related questions, achieving the state-of-the-art performance (14.2\% improvement over the previous SOTA) on the latest EHR-QA dataset (\textit{i.e.} MIMICSQL*).
    \item We propose a simple and effective training objective, Input-Masking (IM) which is an effective solution to cope with various input typos. We demonstrate that our masking strategy yields about 28.8\% improvement over the previous state-of-the-art model for typo-ridden MIMICSQL*.
    \item We further conducted a comprehensive analysis and confirmed the efficacy of \model, especially verifying the consistent results on the latest graph-based EHR-QA dataset (\textit{i.e.} MIMICSPARQL*)
\end{itemize}

\section{Related Work}
\label{sec:related}
\subsection{NLQ to Query Language Generation}
For question answering over relational databases, translating natural language questions into corresponding queries (NLQ2Query) has become the dominant approach.
Existing NLQ2Query datasets and approaches can be categorized depending on their main purposes: generalization over cross domains or specialization within a target domain.
For the former case, a variety of methods \citep{guo2019towards, choi2020ryansql, wang2020rat} were designed to handle unseen queries or databases during evaluation, mainly based on WikiSQL \citep{zhong2017seq2sql} and Spider \citep{yu2018spider}.
Those models are more focused on inferring complex query structures (\textit{i.e.} SQL syntax, tables, and column names) rather than parsing and predicting desired condition values related to the validity of final execution.
On the other hand, domain-specific datasets \citep{price1990evaluation, quirk2015language, li2014constructing} have been studied for a longer period of time \citep{giordani-moschitti-2012-translating, dong2018coarse}.
Most of datasets, however, are small ($<$ 1,000 samples) and are used as another sources to measure generalization.
For the healthcare domain, there are two publicly available NLQ2Query datasets over EHR database \citep{wang2020text, park2021knowledge} in terms of viewing EHR as relational tables or a massive knowledge graph.

\subsection{Electronic Health Records QA}
Recent EHR-QA research can be classified into two main categories: unstructured QA and structured QA.
For the former case, QA research has been mainly developed as machine reading comprehension which extracts the answer to the given question from free text such as clinical case reports \citep{vsuster2018clicr}, clinical notes \citep{pampari2018emrqa}, and healthcare articles \citep{zhu2020question}.

For the structured case, it can be divided into table-based QA and graph-based QA according to the structure of the knowledge base.
TREQS \citep{wang2020text} is the table-based QA model solving the NLQ2SQL task over MIMICSQL which is an EHR-QA dataset derived from MIMIC-III~\citep{johnson2016mimic}, an open-source dataset for ICU records.
\cite{raghavan2021emrkbqa} also proposed emrKBQA, but not publicly available yet, another table-based QA dataset aimed at semantic parsing to map natural language questions to logical forms from the structural part of the EHR (\textit{i.e.} MIMIC-III).
For the counterpart of table-based QA, \cite{park2021knowledge} extended the field of structured EHR-QA by transforming MIMICSQL's tables to a knowledge graph, thus proposing a graph-based EHR-QA.
Furthermore, \cite{park2021knowledge} empirically demonstrated that a graph-based approach is more suitable for conducting complex EHR-QA than a table-based approach.
In all previous works, however, the EHR-QA task was tackled with the classical encoder-to-decoder architecture implemented with RNNs, without considering the complex nature of EHR and the practical challenges induced by it.

\vspace{-3mm}
\section{Method}
\label{sec:method}
\vspace{-1mm}
\subsection{Problem Setup}
\vspace{-1mm}
Our goal is to transform the natural questions asked by the user into executable queries (\textit{i.e.} SQL or SPARQL).
For notation, we define a natural language question $Q$ as a series of $n$ tokens (\textit{i.e.} subwords), $Q=\{q_1, q_2, \ldots, q_n\}$.
Similarly, we define $Y$ as the corresponding query consisting of $m$ tokens, $Y=\{y_1, y_2, \ldots, y_m\}$.
The goal of our model is to maximize the probability $P(Y|Q)$.
Note that similar to \cite{wang2020text} and \cite{park2021knowledge}, we do not rely on the meta information of the knowledge source (\textit{e.g.} database schema or knowledge graph structure) in order to improve the generality of our approach (\textit{i.e.} usable for both NLQ2SQL and NLQ2SPARQL).
Instead, we assume that the meta information is implicitly expressed by the natural language questions, and let the model learn it via the training process.

\begin{figure}[t]
\floatconts
    {fig:model}
    {\vspace{-5mm}\caption{Overview of training our \model model. The input $q_i$ and $y_j$ are summed with the corresponding segment embedding $s_{0/1}$ and the position embedding $p_i, p'_j$, respectively. Two training objectives are used: Masked LM for the question part and seq-to-seq LM for the query language part. The colored line represents whether a pair of tokens can be attended to each other.}\vspace{-1em}}
    {\includegraphics[width=1.0\columnwidth]{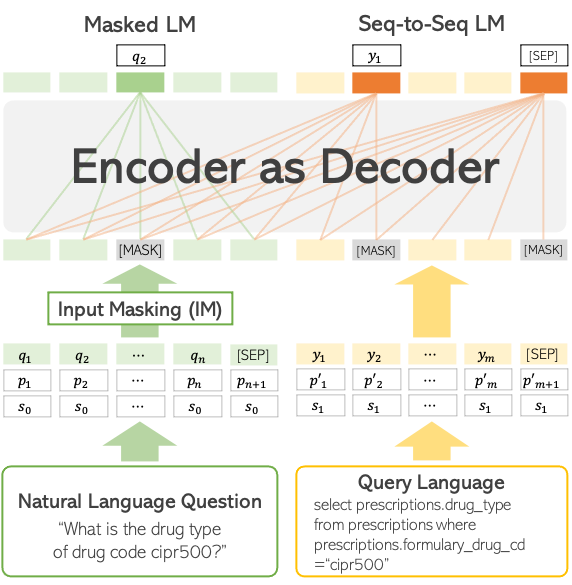}}
\end{figure}

\vspace{-3mm}
\subsection{Input Representation}
\vspace{-1mm}
\label{sssec:input}
Let the model input be a sequence which consists of two sub-sequences: the natural language question $Q=\{q_1, q_2, \ldots, q_n\}$ and corresponding query $Y=\{y_1, y_2, \ldots, y_m\}$.
Input questions and queries consist of multiple tokens, tokenized by subword units by WordPiece \citep{wu2016google} regardless of sub-sequence types.
Both $Q$ and $Y$ are converted to a sequence of token embeddings $\{\boldsymbol{q}_1, \boldsymbol{q}_2, \ldots, \boldsymbol{q}_n\}$ and $\{\yb_1, \yb_2, \ldots, \yb_m\}$ via a trainable lookup table, resepectively.
Then we place a special separator token embedding $[SEP]$ each at the end of $Q$ and $Y$ to form a model input as shown in Figure ~\ref{fig:model}.
Then each token embedding is summed with the corresponding position embedding $\boldsymbol{p}$ and segment embedding $\boldsymbol{s}$ to prepare the final input to the model. 

\subsection{Encoder-as-Decoder Architecture}
Inspired by \cite{dong2019unified}, we propose an \textit{Encoder-as-Decoder} model suited for the NLQ2Query task.
To the best of our knowledge, this is the first attempt to adapt the encoder-as-decoder framework into the NLQ2Query task.
Denoting the input embeddings from Section~\ref{sssec:input} as $\boldsymbol{H}^{0}$, they are encoded into contextual representations at different levels of hidden outputs $\boldsymbol{H}^{l}$ using an $L$-layer Transformer encoders $\boldsymbol{H}^l=\text{Transformer}(\boldsymbol{H}^{l-1}), \; l\in[1,L]$ \citep{vaswani2017attention}.

In this unified architecture, decoding is performed similar to \cite{dong2019unified};
At inference, given the input sequence ($Q$, [SEP], [MASK]$_1$), the model predicts  $\hat{y}_1$, the identity of [MASK]$_1$. Then [MASK]$_1$ is replaced with $\hat{y}_1$ and we attach [MASK]$_2$ to the previous input sequence and repeat this process until [SEP] token is predicted.

In a typical \textit{Encoder-to-Decoder} architecture, the decoder can only access the fully contextualized input embeddings (\textit{i.e.} the output of the encoder), thereby limiting the model's ability to consider the input tokens during the decoding process. 
On the other hand, the \textit{Encoder-as-Decoder} architecture allows the decoding process to access the input tokens at every layer of the encoder, thus improving the decoding capacity.
Moreover, thanks to both encoding and decoding trained with [MASK] reconstruction (unlike Encoder-to-Decoder where the decoder is trained autoregressively), Encoder-as-Decoder can be naturally initialized with pre-trained language models such as BERT~\citep{devlin2019bert}.

\vspace{-1mm}
\subsection{Input Masking on NLQ}
Natural questions in the EHR-QA are prone to various typos such as reversed or missing letters, due to the complex and specialized medical terms.
In order to make the encoder-as-decoder model more robust to various typos, we use an additional training strategy named \textit{Input Masking}.
During training, tokens in the input NLQ $\{q_1, q_2, \ldots, q_n\}$ are randomly masked with probability 0.2. We replace the chosen token with the [MASK] token with 80\% chance, a random token with 10\% chance, and the original token with 10\% chance.
This technique, which can be seen as a form of data augmentation, encourages the model to rely more on familiar tokens when unfamiliar tokens (\textit{i.e.} typos) are included in the input.

In addition, Input Masking can help the model better learn the syntactic structure of SQL (or SPARQL) queries.
For example, typos can occur at various places other than medical terms in the NLQ (\textit{i.e.} retrieve can be misspelled as retreive), which can obstruct accurate decoding.
IM, however, can alleviate this type of challenge as it will make the model robust to any kind of typos in the NLQ.
We empirically demonstrate the effectiveness of Input Masking in the experiments by evaluating the proposed training strategy on datasets with different levels of noise (\textit{i.e.} typos).

\vspace{-1mm}
\subsection{Model Training}
We initialize the encoder-as-decoder model with BERT-base~\citep{devlin2019bert} and train our model with the two training objectives: 1) Masked Language Modeling (MLM) is used for reconstructing the input masks applied to the NLQ part;
2) Sequence-to-Sequence (Seq2Seq) is used to train our model to act as a decoder. 
The difference between MLM and Seq2Seq is the structure of the attention masks.
As shown in Fig.~\ref{fig:model}, the NLQ tokens can freely attend to one another but not to the query tokens.
The query tokens, on the other hand, can only attend to the previous query tokens but freely attend to all NLQ tokens.
By jointly optimizing with two aforementioned training objectives, the model should recover for masked NLQ tokens while also predicting the corresponding query tokens.

\vspace{-2mm}
\section{Experiments}
\label{sec:results}
\subsection{Experiment Settings}

\subsubsection{Dataset}
\textbf{MIMICSQL*}
For evaluating the models, we use MIMICSQL*~\citep{park2021knowledge}, which is a table-based EHR-QA dataset consisting of $10,000$ NLQ-SQL pairs derived from 9 tables\footnote{Patients, Admissions, Diagnoses, Prescriptions, Procedures, Lab Results, Diagnosis Code Dictionary, Procedure Code Dictionary, Lab Code Dictionary} of MIMIC-III~\citep{johnson2016mimic}, an open-source ICU dataset.
For all experiments, we use 8,000 pairs for training, 1,000 for validation and 1,000 for testing.
Note that MIMICSQL* is a revised version of MIMISQL~\citep{wang2020text}, where the authors restored the original MIMIC-III schema as well as improved the tokenization of both NLQ and SQL. \\[3pt]
\textbf{NOISY MIMICSQL*}
As shown in Figure~\ref{fig:nlq_length}, MIMICSQL* has longer input questions on average compared to existing NLQ2Query datasets, which can cause users to make typographical errors more often than the existing datasets.
Therefore, in order to simulate such realistic scenarios, we created three MIMICSQL* variants, modified with different levels of NLQ typos: noise-\textit{weak}, noise-\textit{moderate}, and noise-\textit{strong}.
The details about the noise generation process are presented in the following section.

\begin{figure}[t]
\floatconts
    {fig:nlq_length}
    {\caption{The distributions of the character-level NLQ length over various domain-specific NLQ2SQL datasets.}\vspace{-2em}}
    {\includegraphics[width=\linewidth]{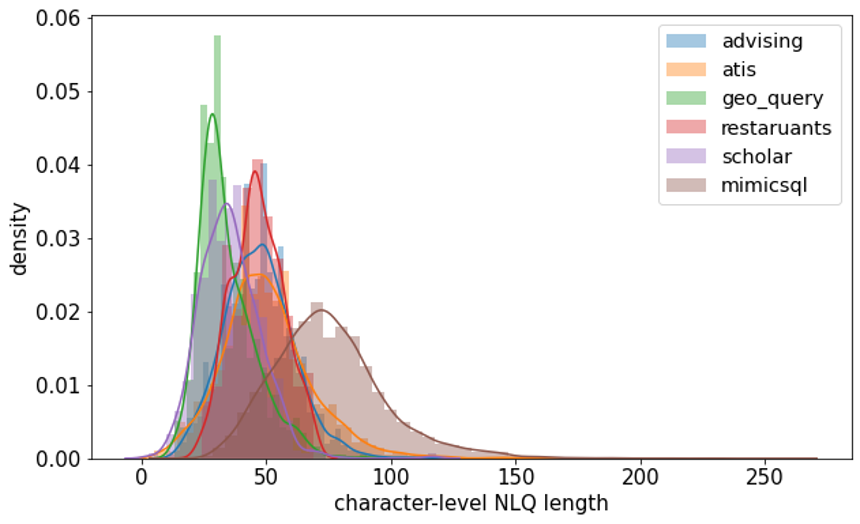}\vspace{-2em}}
\end{figure}

\vspace{-1mm}
\subsubsection{Noise Generation}
\label{sssec:noise_gen}
Following \cite{kemighan1990spelling}, we adopt four different types of common typos: Reversal, Substitution, Deletion, and Insertion.
When we corrupt a single word, we use one of the four types with the fixed ratio of 50\%, 20\%, 15\%, and 15\%, respectively.
The detailed descriptions of four types are as follows:
\begin{itemize}[noitemsep,topsep=0pt]
    \item Reversal: Flip two adjacent characters in a word, such as ``number'' $\rightarrow$ ``numbre''.
    \item Substitution: Mistype a nearby key on a keyboard, such as ``diagnosis'' $\rightarrow$ ``diagnowis''.
    \item Deletion: Delete a random character of the word, such as ``acetylcysteine'' $\rightarrow$ ``acetylcyseine''.
    \item Insertion: Insert a character into the word, such as ``coronary'' $\rightarrow$  ``copronary''.
\end{itemize}

Based on the four pre-defined types of typos and their fixed ratios, we corrupt the target dataset with our noise generator via the following process.
First, given a single natural language question, we split the sentence into individual words. 
For each word, the noise generator determines whether to make a typo based on its length and uniformly sampled probability $p$.
If the generator determines to create a typo for that word, one of the four typo types are applied.
The noise generator repeats this process for all NLQs in the target dataset.

We also make heuristic rules to prevent a sentence or its specific values from losing its original meaning due to typos.
1) The date-time value (e.g. ``16:00:00'') should not be changed;
2) The numerical value (e.g. ``47'' for age value) should not be changed;
3) The short word, smaller than the minimum length hyperparameter ($l_\textsf{min}$), should not be changed.

By using our noise generator, we apply three different degrees of noise (noise-\textit{weak}, noise-\textit{moderate}, and noise-\textit{strong}) to MIMICSQL*, and use them only at the evaluation phase. 
Note that noise-\textit{weak} contains approximately 5\% of corrupted words in a sentence on average, noise-\textit{moderate} about 10\%, and noise-\textit{strong} about 15\%.
The detailed algorithm is presented in the Appendix~\ref{appendix:implement}.

We believe our three-level typo-ridden approach can reasonably represent realistic scenarios on MIMICSQL* for four reasons: 1) \cite{cucerzan-brill-2004-spelling} reports that users make 10-15\% spelling errors in their queries when using search engines;
2) Following \cite{hagiwara-mita-2020-github}, we confirm that real-world typo datasets have misspellings of approximately 10\%;
3) MIMICSQL* has the longest natural language question compared to existing domain-specific NLQ2SQL datasets, due to the complex medical terms, as shown in Figure~\ref{fig:nlq_length};
4) We found roughly a dozen instances with real-world typos in the MIMICSQL* \textit{test} dataset, despite the fact the dataset went through manual inspection during construction. 
Also, we observed all their typos (\textit{e.g.} ``ethnicty'', ``nymber'') occur at the character-level, confirming the realism of our synthetic noise injection method.


\vspace{-2mm}
\subsubsection{Comparison Methods}
\vspace{-1mm}
We compare our model (\textbf{\model}) with the following baseline models.
In all experiments, all models were trained with five random seeds, and we report the mean and the standard deviation.
Further implementation and hyperparameter details are provided in the Appendix~\ref{appendix:implement}. \\[3pt]
\textbf{Seq2Seq + Attention}
Seq2Seq with attention \citep{luong2015effective} consists of a bidirectional LSTM encoder and an LSTM decoder. Following the original paper, we apply the attention mechanism in this model. It should be noted this model cannot handle the out-of-vocabulary (OOV) tokens. We simply denote the model as Seq2Seq. \\[3pt]
\textbf{TREQS}
TREQS \citep{wang2020text} is the state-of-the-art NLQ2SQL model on MIMICSQL*. This LSTM-based encoder-to-decoder model uses temporal attention on NLQ, dynamic attention on SQL to capture the condition values accurately, and copy-mechanism to resolve the OOV problem. \\[3pt]
\textbf{E-to-D}
TREQS is also an encoder-to-decoder model, but it does not use self-attention layers, and the model size is not comparable to ours. 
Accordingly, we adopt BERT2BERT \citep{rothe2020leveraging} to utilize the pre-trained BERT(6-layer, 768-hidden, 12-head) in both the encoder and the decoder, making the number of model parameters similar to \model.
Its structure is exactly the same as the Transformer \citep{vaswani2017attention}. 
\\[3pt]
\textbf{E-to-D + IM} To provide a fair opportunity for the E-to-D model with the input masking strategy (IM), we trained the E-to-D model with masked language modeling on the encoder side while using the autoregressive modeling on the decoder side. \\[3pt]
\textbf{E-as-D}
In order to verify the effectiveness of the IM strategy, we use the vanilla encoder-as-decoder model, initialized with BERT(12-layer, 768-hidden, 12-head).

\vspace{-1mm}
\subsubsection{Evaluation Metrics}
\vspace{-1mm}
To evaluate the QA performance of different NLQ2Query models, we use the same evaluation metrics as described in previous works \citep{wang2020text, park2021knowledge}.
1) \textit{Logical Form Accuracy} ($Acc_{LF}$) is computed by comparing the generated SQL/SPARQL queries with the true SQL/SPARQL queries token-by-token;
2) \textit{Execution Accuracy} ($Acc_{EX}$) represents the matching ratio between results from executing the generated query and the results from executing the ground truth query.
Note that it is possible for incorrect queries to luckily return correct results and affect this metric (\textit{e.g.} when the target result is 0 or Null); 
3) \textit{Structural Accuracy ($Acc_{ST}$)} is equivalent to $Acc_{LF}$ except that the condition value tokens (e.g. numeric values or string values) are ignored, therefore focusing on the SQL/SPARQL syntactic structure only. 

\vspace{-1mm}
\subsubsection{Recovering for condition values}
\vspace{-1mm}
In addition to Input Masking, we use the condition value recovery technique used in \cite{wang2020text} to better handle condition values often containing complex medical terms.
After the NLQ2Query model generates a query, this technique is used to compare the condition values in the generated query to the existing values in the database.
Then, the condition values in the generated query are replaced with the most similar (or identical) values in the database.
For example, if the user asks for the number of patients with \textit{essential hypertension}, then the generated query will contain the condition value \textit{essential hypertension}.
But if this value does not exist in the database, the recovery technique will calculate the ROUGE-L scores between \textit{essential hypertension} with all values in the database.
Then the closest value, for example \textit{essential hypertensive disorder}, is chosen and replaced with \textit{essential hypertension}, thus making the generated query executable.

Note that both the recovery technique and the proposed IM training strategy have a similar purpose, namely handling complex medical terms.
However, the recovery technique is a post-processing technique for the generated SQL/SPARQL query.
If the generated query is an incorrect query to begin with due to noisy NLQ, the recovery technique would only rectify the condition values, but the entire query would still be incorrect.
From this perspective, another role of IM, which not only adjusts typos but also captures query structure, can be combined with this recovery technique, where both techniques can complement each other effectively.
We will further discuss this in the following experiment and analysis section.





\begin{table}[t]
\floatconts
    {tab:base_exp}
    {\vspace{-1.5em}\caption{
    Test-set results on MIMICSQL*. We report the mean and standard deviation of the three evaluation metrics ($Acc_{LF}$, $Acc_{EX}$, and $Acc_{ST}$) over 5 random seeds.
    Note that E-to-D=Encoder-to-Decoder, E-as-D=Encoder-as-Decoder, IM=Input Masking.}\vspace{-2em}}%
    {%
        \resizebox{\columnwidth}{!}{%
        \begin{tabular}{lllllll}
            \hline
            \textbf{Method}
            & \multicolumn{3}{c}{\textbf{Test Performance for MIMICSQL*}} \\
            &
              \multicolumn{1}{c}{$Acc_{LF}$} &
              \multicolumn{1}{c}{$Acc_{EX}$} &
              \multicolumn{1}{c}{$Acc_{ST}$} \\ \hline
            \textbf{Before Recovering} \\
            Seq2Seq &
              $0.128\ {\scriptstyle (0.077)}$ &
              $0.263\ {\scriptstyle (0.066)}$ &
              $0.338\ {\scriptstyle (0.030)}$ \\
            TREQS &
              $0.604\ {\scriptstyle (0.008)}$ &
              $0.694\ {\scriptstyle (0.004)}$ &
              $0.799\ {\scriptstyle (0.008)}$ \\
            E-to-D &
              $0.832\ {\scriptstyle (0.006)}$ &
              $0.884\ {\scriptstyle (0.005)}$ &
              $0.900\ {\scriptstyle (0.006)}$ \\
            E-to-D + IM &
              $0.790\ {\scriptstyle (0.006)}$ &
              $0.854\ {\scriptstyle (0.002)}$ &
              $0.878\ {\scriptstyle (0.007)}$ \\
            E-as-D &
              $\textbf{0.856}\ {\scriptstyle (0.008)}$ &
              $\textbf{0.900}\ {\scriptstyle (0.007)}$ &
              $0.894\ {\scriptstyle (0.003)}$ \\
            \model (E-as-D + IM) &
              $0.849\ {\scriptstyle (0.011)}$ &
              $0.895\ {\scriptstyle (0.009)}$ &
              $\textbf{0.905}\ {\scriptstyle (0.013)}$ \\ \hline
            \textbf{After Recovering} \\
            Seq2Seq &
              $0.136\ {\scriptstyle (0.081)}$ &
              $0.242\ {\scriptstyle (0.071)}$ &
              $0.338\ {\scriptstyle (0.030)}$ \\
            TREQS &
              $0.740\ {\scriptstyle (0.006)}$ &
              $0.822\ {\scriptstyle (0.009)}$ &
              $0.799\ {\scriptstyle (0.008)}$ \\
            E-to-D &
              $0.867\ {\scriptstyle (0.006)}$ &
              $0.920\ {\scriptstyle (0.008)}$ &
              $0.900\ {\scriptstyle (0.006)}$ \\
            E-to-D + IM &
              $0.792\ {\scriptstyle (0.015)}$ &
              $0.875\ {\scriptstyle (0.004)}$ &
              $0.878\ {\scriptstyle (0.007)}$ \\
            E-as-D &
              $0.878\ {\scriptstyle (0.006)}$ &
              $0.928\ {\scriptstyle (0.004)}$ &
              $0.894\ {\scriptstyle (0.003)}$ \\
            \model (E-as-D + IM) &
              $\textbf{0.882}\ {\scriptstyle (0.012)}$ &
              $\textbf{0.934}\ {\scriptstyle (0.008)}$ &
              $\textbf{0.905}\ {\scriptstyle (0.013)}$ \\ \hline
        \end{tabular}
        }%
    }%
\vspace{-3mm}
\end{table}

\begin{table*}[ht]
\centering
\floatconts
    {tab:noise_exp}
    {\caption{Test-set results on NOISY MIMICSQL*. 
    We report the mean and standard deviation of the three evaluation metrics ($Acc_{LF}$, $Acc_{EX}$, and $Acc_{ST}$) over 5 random seeds.
    Based on the probability of corrupting words on average in a sentence, we refer to 5\% as noise-\textit{weak}, 10\% as noise-\textit{moderate}, and 15\% as noise-\textit{strong}.}}
    {\vspace{-1.5em}}
    {\resizebox{\textwidth}{!}{
    \begin{tabular}{llllllllll}
    \hline
    {\textbf{Method}} &
      \multicolumn{3}{c}{\textbf{noise-\textit{weak} (5\% typo prob.)}} &
      \multicolumn{3}{c}{\textbf{noise-\textit{moderate} (10\% typo prob.)}} &
      \multicolumn{3}{c}{\textbf{nosie-\textit{strong} (15\% typo prob.)}} \\
     &
      \multicolumn{1}{c}{$Acc_{LF}$} &
      \multicolumn{1}{c}{$Acc_{EX}$} &
      \multicolumn{1}{c}{$Acc_{ST}$} &
      \multicolumn{1}{c}{$Acc_{LF}$} &
      \multicolumn{1}{c}{$Acc_{EX}$} &
      \multicolumn{1}{c}{$Acc_{ST}$} &
      \multicolumn{1}{c}{$Acc_{LF}$} &
      \multicolumn{1}{c}{$Acc_{EX}$} &
      \multicolumn{1}{c}{$Acc_{ST}$} \\ \hline
    \textbf{Before Recovering} \\
    Seq2Seq &
      $0.096\ {\scriptstyle (0.062)}$ &
      $0.226\ {\scriptstyle (0.056)}$ &
      $0.264\ {\scriptstyle (0.032)}$ &
      $0.063\ {\scriptstyle (0.033)}$ &
      $0.169\ {\scriptstyle (0.033)}$ &
      $0.173\ {\scriptstyle (0.017)}$ &
      $0.044\ {\scriptstyle (0.024)}$ &
      $0.135\ {\scriptstyle (0.023)}$ &
      $0.119\ {\scriptstyle (0.020)}$ \\
    TREQS &
      $0.418\ {\scriptstyle (0.011)}$ &
      $0.546\ {\scriptstyle (0.010)}$ &
      $0.665\ {\scriptstyle (0.011)}$ &
      $0.281\ {\scriptstyle (0.019)}$ &
      $0.412\ {\scriptstyle (0.024)}$ &
      $0.559\ {\scriptstyle (0.031)}$ &
      $0.228\ {\scriptstyle (0.020)}$ &
      $0.363\ {\scriptstyle (0.024)}$ &
      $0.483\ {\scriptstyle (0.041)}$ \\
    E-to-D &
      $0.709\ {\scriptstyle (0.007)}$ &
      $0.783\ {\scriptstyle (0.006)}$ &
      $0.839\ {\scriptstyle (0.009)}$ &
      $0.554\ {\scriptstyle (0.011)}$ &
      $0.648\ {\scriptstyle (0.013)}$ &
      $0.703\ {\scriptstyle (0.008)}$ &
      $0.456\ {\scriptstyle (0.015)}$ &
      $0.551\ {\scriptstyle (0.016)}$ &
      $0.639\ {\scriptstyle (0.013)}$ \\
    E-to-D + IM &
      $0.665\ {\scriptstyle (0.027)}$ &
      $0.754\ {\scriptstyle (0.025)}$ &
      $0.791\ {\scriptstyle (0.025)}$ &
      $0.490\ {\scriptstyle (0.031)}$ &
      $0.593\ {\scriptstyle (0.027)}$ &
      $0.638\ {\scriptstyle (0.036)}$ &
      $0.417\ {\scriptstyle (0.032)}$ &
      $0.522\ {\scriptstyle (0.031)}$ &
      $0.576\ {\scriptstyle (0.032)}$ \\
    E-as-D &
      $0.654\ {\scriptstyle (0.005)}$ &
      $0.738\ {\scriptstyle (0.005)}$ &
      $0.850\ {\scriptstyle (0.004)}$ &
      $0.510\ {\scriptstyle (0.016)}$ &
      $0.615\ {\scriptstyle (0.017)}$ &
      $0.749\ {\scriptstyle (0.016)}$ &
      $0.382\ {\scriptstyle (0.018)}$ &
      $0.493\ {\scriptstyle (0.016)}$ &
      $0.684\ {\scriptstyle (0.025)}$ \\
    UniQA (E-as-D + IM) &
      $\textbf{0.740}\ {\scriptstyle (0.027)}$ &
      $\textbf{0.807}\ {\scriptstyle (0.021)}$ &
      $\textbf{0.877}\ {\scriptstyle (0.016)}$ &
      $\textbf{0.610}\ {\scriptstyle (0.035)}$ &
      $\textbf{0.691}\ {\scriptstyle (0.028)}$ &
      $\textbf{0.784}\ {\scriptstyle (0.028)}$ &
      $\textbf{0.504}\ {\scriptstyle (0.032)}$ &
      $\textbf{0.602}\ {\scriptstyle (0.028)}$ &
      $\textbf{0.731}\ {\scriptstyle (0.024)}$ \\ \hline
    \textbf{After Recovering} \\
    Seq2Seq &
      $0.103\ {\scriptstyle (0.065)}$ &
      $0.201\ {\scriptstyle (0.059)}$ &
      $0.264\ {\scriptstyle (0.032)}$ &
      $0.066\ {\scriptstyle (0.035)}$ &
      $0.141\ {\scriptstyle (0.036)}$ &
      $0.173\ {\scriptstyle (0.017)}$ &
      $0.046\ {\scriptstyle (0.027)}$ &
      $0.105\ {\scriptstyle (0.026)}$ &
      $0.119\ {\scriptstyle (0.020)}$ \\
    TREQS &
      $0.545\ {\scriptstyle (0.012)}$ &
      $0.668\ {\scriptstyle (0.009)}$ &
      $0.665\ {\scriptstyle (0.011)}$ &
      $0.397\ {\scriptstyle (0.025)}$ &
      $0.531\ {\scriptstyle (0.029)}$ &
      $0.559\ {\scriptstyle (0.031)}$ &
      $0.325\ {\scriptstyle (0.032)}$ &
      $0.464\ {\scriptstyle (0.033)}$ &
      $0.483\ {\scriptstyle (0.041)}$ \\ 
    E-to-D &
      $0.783\ {\scriptstyle (0.011)}$ &
      $0.848\ {\scriptstyle (0.010)}$ &
      $0.839\ {\scriptstyle (0.009)}$ &
      $0.628\ {\scriptstyle (0.012)}$ &
      $0.714\ {\scriptstyle (0.015)}$ &
      $0.703\ {\scriptstyle (0.008)}$ &
      $0.533\ {\scriptstyle (0.013)}$ &
      $0.629\ {\scriptstyle (0.012)}$ &
      $0.639\ {\scriptstyle (0.013)}$ \\
    E-to-D + IM &
      $0.681\ {\scriptstyle (0.028)}$ &
      $0.773\ {\scriptstyle (0.021)}$ &
      $0.791\ {\scriptstyle (0.025)}$ &
      $0.505\ {\scriptstyle (0.030)}$ &
      $0.610\ {\scriptstyle (0.027)}$ &
      $0.638\ {\scriptstyle (0.036)}$ &
      $0.433\ {\scriptstyle (0.032)}$ &
      $0.546\ {\scriptstyle (0.033)}$ &
      $0.576\ {\scriptstyle (0.032)}$ \\
    E-as-D &
      $0.818\ {\scriptstyle (0.006)}$ &
      $0.880\ {\scriptstyle (0.004)}$ &
      $0.850\ {\scriptstyle (0.004)}$ &
      $0.692\ {\scriptstyle (0.014)}$ &
      $0.767\ {\scriptstyle (0.015)}$ &
      $0.749\ {\scriptstyle (0.016)}$ &
      $0.602\ {\scriptstyle (0.021)}$ &
      $0.672\ {\scriptstyle (0.020)}$ &
      $0.668\ {\scriptstyle (0.050)}$ \\
    UniQA (E-as-D + IM) &
      $\textbf{0.825}\ {\scriptstyle (0.018)}$ &
      $\textbf{0.895}\ {\scriptstyle (0.013)}$ &
      $\textbf{0.877}\ {\scriptstyle (0.016)}$ &
      $\textbf{0.695}\ {\scriptstyle (0.028)}$ &
      $\textbf{0.775}\ {\scriptstyle (0.023)}$ &
      $\textbf{0.784}\ {\scriptstyle (0.028)}$ &
      $\textbf{0.612}\ {\scriptstyle (0.021)}$ &
      $\textbf{0.709}\ {\scriptstyle (0.018)}$ &
      $\textbf{0.731}\ {\scriptstyle (0.024)}$ \\ \hline
    \end{tabular}}}
\vspace{-3mm}
\end{table*}

\subsection{Experiments Results}
\vspace{-1mm}
\subsubsection{Results from MIMICSQL*}
\vspace{-1mm}
Model performance on MIMICSQL* test set are shown in Table \ref{tab:base_exp}.
After applying the recovering technique for the condition value, we can observe that \model outperforms the previous state-of-the-art model TREQS by a significant margin 14.2\%.
Thanks to IM, by simply applying the input masking to the vanilla E-as-D model, we can observe that our model has the best structural accuracy over all baselines. 
We also confirm that the performance gain is higher than E-as-D in logical form accuracy compared to before applying the recovering technique, indicating that the recovering technique (\textit{i.e.} replace with proper condition value after the decoding stage) and IM's ability (\textit{i.e.} address typos preemptively at the decoding stage, and capture the SQL syntax) worked in harmony effectively.



\begin{table}[t]
\vspace{-5mm}
\floatconts
{tab:base_sparql_exp}
{\caption{
Test-results on MIMICSPARQL*.
We report the mean and standard deviation of the three evaluation metrics ($Acc_{LF}$, $Acc_{EX}$, and $Acc_{ST}$) over 5 random seeds.
}}
{
    \resizebox{\columnwidth}{!}{%
    \begin{tabular}{lllllll}
    \hline
    \textbf{Method}
    & \multicolumn{3}{c}{\textbf{Test Performance for MIMICSPARQL*}} \\
    &
      \multicolumn{1}{c}{$Acc_{LF}$} &
      \multicolumn{1}{c}{$Acc_{EX}$} &
      \multicolumn{1}{c}{$Acc_{ST}$} \\ \hline
    \textbf{Before Recovering} \\
    Seq2Seq &
      $0.219\ {\scriptstyle (0.053)}$ &
      $0.336\ {\scriptstyle (0.042)}$ &
      $0.339\ {\scriptstyle (0.010)}$ \\
    TREQS &
      $0.603\ {\scriptstyle (0.020)}$ &
      $0.702\ {\scriptstyle (0.014)}$ &
      $0.761\ {\scriptstyle (0.021)}$ \\
    E-to-D &
      $0.823\ {\scriptstyle (0.006)}$ &
      $0.875\ {\scriptstyle (0.004)}$ &
      $0.890\ {\scriptstyle (0.003)}$ \\
    E-to-D + IM &
      $0.776\ {\scriptstyle (0.008)}$ &
      $0.842\ {\scriptstyle (0.011)}$ &
      $0.864\ {\scriptstyle (0.006)}$ \\
    E-as-D &
      $\textbf{0.866}\ {\scriptstyle (0.005)}$ &
      $\textbf{0.909}\ {\scriptstyle (0.004)}$ &
      $0.903\ {\scriptstyle (0.002)}$ \\
    \model (E-as-D + IM) &
      $0.853\ {\scriptstyle (0.009)}$ &
      $0.897\ {\scriptstyle (0.01)}$ &
      $\textbf{0.909}\ {\scriptstyle (0.011)}$ \\ \hline
    \textbf{After Recovering} \\
    Seq2Seq &
      $0.222\ {\scriptstyle (0.055)}$ &
      $0.346\ {\scriptstyle (0.044)}$ &
      $0.339\ {\scriptstyle (0.010)}$ \\
    TREQS &
      $0.610\ {\scriptstyle (0.022)}$ &
      $0.715\ {\scriptstyle (0.012)}$ &
      $0.761\ {\scriptstyle (0.021)}$ \\
    E-to-D &
      $0.864\ {\scriptstyle (0.005)}$ &
      $0.912\ {\scriptstyle (0.004)}$ &
      $0.890\ {\scriptstyle (0.003)}$ \\
    E-to-D + IM &
      $0.792\ {\scriptstyle (0.007)}$ &
      $0.860\ {\scriptstyle (0.008)}$ &
      $0.864\ {\scriptstyle (0.006)}$ \\
    E-as-D &
      $\textbf{0.896}\ {\scriptstyle (0.003)}$ &
      $\textbf{0.937}\ {\scriptstyle (0.006)}$ &
      $0.903\ {\scriptstyle (0.002)}$ \\
    \model (E-as-D + IM) &
      $0.891\ {\scriptstyle (0.008)}$ &
      $0.935\ {\scriptstyle (0.005)}$ &
      $\textbf{0.909}\ {\scriptstyle (0.011)}$ \\ \hline
    \end{tabular}%
    }%
}
\end{table}

\subsubsection{Results from Noisy MIMICSQL*}
\vspace{-1mm}
Model performance for noisy MIMICSQL* are shown in Table \ref{tab:noise_exp}.
As the probability of typos rises from weak to strong, we observe that the performances of all models decrease across three evaluation metrics.
Along with our input masking strategy, we can observe \model outperforms all baseline models, especially outperforming the previous state-of-the-art model TREQS by a significant margin 25.8\%.
Furthermore, \model always recorded the best structural accuracy regardless of the degree of typos. This means that the additional post-processing by the end-user (\textit{e.g.} manually typing in correct condition values) has the potential for further increasing the model performance, since structural accuracy is the upper bound of logical form accuracy. 



\begin{table}[t]
\vspace{-1mm}
\centering
\floatconts
    {tab:noise_sparql_exp_moderate}
    {\caption{
    \vspace{-4mm}
    Test-results on NOISY MIMICSPARQL* with the degree of noise-\textit{moderate}.
    We report the mean and standard deviation of the three evaluation metrics ($Acc_{LF}$, $Acc_{EX}$, and $Acc_{ST}$) over 5 random seeds.
    }\vspace{-1.2em}}
    {
    \resizebox{0.95\columnwidth}{!}{%
    \begin{tabular}{lllllll}
    \hline
    \textbf{Method}
    & \multicolumn{3}{c}{\textbf{noise-\textit{moderate} (10\% typo prob.)}} \\
    &
      \multicolumn{1}{c}{$Acc_{LF}$} &
      \multicolumn{1}{c}{$Acc_{EX}$} &
      \multicolumn{1}{c}{$Acc_{ST}$} \\ \hline
    \textbf{Before Recovering} \\
    Seq2Seq &
      $0.093\ {\scriptstyle (0.016)}$ &
      $0.178\ {\scriptstyle (0.019)}$ &
      $0.178\ {\scriptstyle (0.019)}$ \\
    TREQS &
      $0.277\ {\scriptstyle (0.016)}$ &
      $0.426\ {\scriptstyle (0.009)}$ &
      $0.494\ {\scriptstyle (0.019)}$ \\
    E-to-D &
      $0.547\ {\scriptstyle (0.009)}$ &
      $0.639\ {\scriptstyle (0.010)}$ &
      $0.704\ {\scriptstyle (0.016)}$ \\
    E-to-D + IM &
      $0.534\ {\scriptstyle (0.025)}$ &
      $0.631\ {\scriptstyle (0.031)}$ &
      $0.678\ {\scriptstyle (0.032)}$ \\
    E-as-D &
      $0.504\ {\scriptstyle (0.006)}$ &
      $0.609\ {\scriptstyle (0.010)}$ &
      $0.715\ {\scriptstyle (0.024)}$ \\
    \model (E-as-D + IM) &
      $\textbf{0.646}\ {\scriptstyle (0.015)}$ &
      $\textbf{0.725}\ {\scriptstyle (0.015)}$ &
      $\textbf{0.798}\ {\scriptstyle (0.016)}$ \\ \hline
    \textbf{After Recovering} \\
    Seq2Seq &
      $0.094\ {\scriptstyle (0.017)}$ &
      $0.182\ {\scriptstyle (0.020)}$ &
      $0.178\ {\scriptstyle (0.019)}$ \\
    TREQS &
      $0.283\ {\scriptstyle (0.017)}$ &
      $0.434\ {\scriptstyle (0.008)}$ &
      $0.494\ {\scriptstyle (0.019)}$ \\
    E-to-D &
      $0.618\ {\scriptstyle (0.017)}$ &
      $0.708\ {\scriptstyle (0.021)}$ &
      $0.704\ {\scriptstyle (0.016)}$ \\
    E-to-D + IM &
      $0.546\ {\scriptstyle (0.024)}$ &
      $0.644\ {\scriptstyle (0.029)}$ &
      $0.678\ {\scriptstyle (0.032)}$ \\
    E-as-D &
      $0.668\ {\scriptstyle (0.018)}$ &
      $0.740\ {\scriptstyle (0.022)}$ &
      $0.715\ {\scriptstyle (0.024)}$ \\
    \model (E-as-D + IM) &
      $\textbf{0.725}\ {\scriptstyle (0.010)}$ &
      $\textbf{0.799}\ {\scriptstyle (0.011)}$ &
      $\textbf{0.798}\ {\scriptstyle (0.016)}$ \\ \hline
    \end{tabular}%
    }%
}
\vspace{-5mm}
\end{table}

\section{Discussion}
\vspace{-1mm}
\label{sec:analysis}

\begin{figure*}[t]
\vspace{-5mm}
\floatconts
    {fig:noise_qual}
    {\caption{
    \vspace{-5mm}
    SQL Queries generated by different models given the same noisy NLQ.
    The typos in the NLQs are marked in bold, and incorrectly predicted tokens in the generated SQL queries are highlighted in red.
    The correctly predicted SQL tokens, associated with the typo in the NLQ, are highlighted in blue.
    ($\dagger$) denotes a sample that can be corrected by the recovering technique.
    Note that the recovering technique can help only one case out of 10 incorrectly generated queries, indicating the effectiveness of the input masking strategy for noisy NLQs.
    }
    
    }
    {
        \resizebox{\textwidth}{!}{
        \begin{tabular}{p{2.5cm}|p{12.5cm}|p{11.5cm}}
        \toprule
        noisy NLQ
        &   \text{how many patients were given the drug \textbf{ferros} gluconate?} \newline
            \text{(typo: ferrous $\rightarrow$ ferros)}
        &   \text{which \textbf{laguage} does cynthia gomez understand?} \newline
            \text{(typo: language $\rightarrow$ laguage)}
        \\ \midrule
        Ground \newline Truth
        & \text{select count ( distinct patients.subject\_id ) from patients} \newline
        \text{inner join admissions on patients.subject\_id = admissions.subject\_id} \newline
        \text{inner join prescriptions on admissions.hadm\_id = prescriptions.hadm\_id} \newline
        \text{where prescriptions.drug = ``ferrous gluconate''}
        &   \text{select admissions.language from admissions} \newline
            \text{inner join patients on admissions.subject\_id = patients.subject\_id} \newline
            \text{where patients.name = ``cynthia gomez''}
        \\ \hline
        Seq2Seq
        &   \text{select count ( distinct patients.subject\_id ) from patients} \newline
            \text{inner join admissions on patients.subject\_id = admissions.subject\_id} \newline
            \text{inner join prescriptions on admissions.hadm\_id = prescriptions.hadm\_id} \newline
            \text{where prescriptions.drug = \textcolor{red}{$<$UNK$>$ $<$UNK$>$}}
        &   \text{select \textcolor{red}{d\_icd\_diagnoses.short\_title} from \textcolor{red}{d\_icd\_diagnoses}} \newline
            \text{inner join \textcolor{red}{diagnoses} on \textcolor{red}{d\_icd\_diagnoses.icd9\_code = diagnoses.icd9\_code}} \newline
            \text{where patients.name = \textcolor{red}{$<$UNK$>$ $<$UNK$>$}}
        \\ \hline
        TREQS
        &   \text{select count ( distinct patients.subject\_id ) from patients} \newline
            \text{inner join admissions on patients.subject\_id= admissions.subject\_id} \newline
            \text{inner join prescriptions on admissions.hadm\_id=prescriptions.hadm\_id} \newline
            \text{where \textcolor{red}{admissions.age $<$ ``44'' and prescriptions.drug = ``glucohate glucohate''}}
        &   \text{select \textcolor{red}{count ( distinct patients.subject\_id)} from patients} \newline
            \text{inner join admissions on patients.subject\_id = admissions.subject\_id} \newline
            \text{\textcolor{red}{inner join prescriptions on admissions.hadm\_id = prescriptions.hadm\_id}} \newline
            \text{\textcolor{red}{where prescriptions.drug = ``gomez sibject''}}
        \\ \hline
        E-to-D
        &   \text{select count ( distinct patients.subject\_id ) from patients} \newline
            \text{inner join admissions on patients.subject\_id = admissions.subject\_id} \newline
            \text{inner join prescriptions on admissions.hadm\_id = prescriptions.hadm\_id} \newline
            \text{where prescriptions.drug = ``\textcolor{red}{ferrous sulfate}''}
        &   \text{select admissions.\textcolor{red}{age} from admissions} \newline
            \text{inner join patients on admissions.subject\_id = patients.subject\_id} \newline
            \text{where patients.name = ``cynthia gomez''}
        \\ \hline
        E-to-D$+$IM 
        &   \text{select count ( distinct patients.subject\_id ) from patients} \newline 
            \text{inner join admissions on patients.subject\_id = admissions.subject\_id} \newline
            \text{inner join prescriptions on admissions.hadm\_id = prescriptions.hadm\_id} \newline
            \text{where prescriptions.drug = ``\textcolor{red}{ferrous sulfate}''}
        &   \text{select admissions.\textcolor{red}{insurance} from admissions} \newline
            \text{inner join patients on admissions.subject\_id = patients.subject\_id} \newline
            \text{where patients.name = ``cynthia gomez''}
        \\ \hline
        E-as-D
        &   \text{select count ( distinct patients.subject\_id ) from patients} \newline
            \text{inner join admissions on patients.subject\_id = admissions.subject\_id} \newline
            \text{inner join prescriptions on admissions.hadm\_id = prescriptions.hadm\_id} \newline
            \text{where prescriptions.drug = ``\textcolor{red}{ferros gluconate}''($\dagger$)}
        &   \text{select admissions.\textcolor{red}{marital\_status} from admissions} \newline
            \text{inner join patients on admissions.subject\_id = patients.subject\_id} \newline
            \text{where patients.name = ``cynthia gomez''}
        \\ \hline
        UniQA \newline (E-as-D + IM)
        &   \text{select count ( distinct patients.subject\_id ) from patients} \newline
            \text{inner join admissions on patients.subject\_id = admissions.subject\_id} \newline
            \text{inner join prescriptions on admissions.hadm\_id = prescriptions.hadm\_id} \newline
            \text{where prescriptions.drug = ``\textcolor{blue}{ferrous gluconate}''}
        &   \text{select admissions.\textcolor{blue}{language} from admissions} \newline
            \text{inner join patients on admissions.subject\_id = patients.subject\_id} \newline
            \text{where patients.name = ``cynthia gomez''}
        \\ \bottomrule
        \end{tabular}}
    }
\vspace{-5mm}
\end{figure*}


\subsection{Graph-based EHR Question Answering}
In terms of viewing EHR as a massive knowledge graph (KG) rather than multiple relational tables, there is another publicly available EHR-QA dataset, MIMICSPARQL*.
Although MIMICSPARQL* has the different style of queries (\textit{i.e.} SPARQL), the answer is eventually the same as MIMICSQL* because it has the same NLQs as MIMICSQL* and tables are transformed into a knowledge graph without any information loss.
To evaluate models over graph-based EHR-QA, we conducted the same experiments as MIMICSQL*, with the original MIMICSPARQL* and its noisy variants.

As shown in Table~\ref{tab:base_sparql_exp} and Table~\ref{tab:noise_sparql_exp_moderate}, we can observe \model shows the consistent results on MIMICSPARQL* and its variants.
As demonstrated in all previous experiments, \model has the highest structural accuracy over all baselines.
In addition, \cite{park2021knowledge} demonstrate that the graph-based approach is more suitable for EHR-QA. 
We also empirically demonstrate that the overall performance of MIMICSPARQL* is higher than MIMICSQL*, consistent with the empirical results from \cite{park2021knowledge}.


\subsection{Why IM does not help in E-to-D case}
As shown in Table~\ref{tab:noise_exp}, the performance of the E-to-D model decreases when the model is combined with the input masking strategy. 
We believe this originates from the architectural difference between E-to-D and E-as-D. In contrast to E-as-D where a single transformer model acts as both encoder and decoder, E-to-D employs two distinct encoder and decoder. Therefore when E-as-D is combined with IM, both encoder and decoder are trained with the reconstruction loss. E-to-D, however, when combined with IM, encoder and decoder are trained with two distinct losses (\textit{i.e.} encoder with reconstruction loss, decoder with auto-regressive loss), which seems to yield a negative effect rather than a synergistic effect.

\subsection{Qualitative Comparison between Generated Queries}
We demonstrate the qualitative results to study differences between models and how each model generates the SQL query given the noisy input question.
As shown in Figure~\ref{fig:noise_qual}, we present the ground truth and generated queries by six models given two NLQs with a typo. 
All results are generated during the evaluation phase on the NOISY MIMICSQL* dataset with the degree of noise-\textit{moderate}.

On the left side, the word \emph{``ferrous gluconate''} in the NLQ, used as the condition value for the SQL query, is corrupted by a single deletion. 
Due to this typo, all models except \model have errors in the condition value part.
Interestingly, we can see that E-to-D generates condition value that is not present in the noisy NLQ, but E-as-D copies the incorrect condition value including a typo.
Thanks to the IM strategy, unlike the vanilla E-as-D model, \model correctly adjusts the corrupted word to its original condition value.

On the right side, the word \emph{``language''} in the NLQ, used as the target column in the SQL query, is also corrupted by a single deletion. 
This simple deletion makes all models except \model generate incorrect SQL queries with incorrect table names or column names.
Here we can confirm that \model can capture table and column names well even when given the noisy input.
Based on the two cases discussed above, it can be seen that \model is robust to the typos regardless of its positions in the input question, namely be it medical terms or other lengthy words.

\subsection{Error Analysis}
To gain intuition and understand the challenging points in EHR-QA, we conducted an error analysis on failure cases (\textit{i.e.} 151 samples) made by \model for MIMICSQL* without any noise.
\paragraph{Insufficient information in the question}
This most popular failure type occurs because the given natural language question provides insufficient or implicit information to generate accurate queries. 
For example, in the question \emph{how many of the male patients had icd9 code 8842?}, the \emph{icd9 code} may refer the \emph{diagnoses\_icd9\_code} or the \emph{procedure\_icd9\_code} column, so the model might generate an incorrect query.
Another example is incorrectly decoding non-specific questions such as \emph{specify details of icd9 code 4591}, which should be decoded to retrieve both short and long names of code 4591, but the model would retrieve only the long name.
If the model can interact with the user (\textit{e.g.} ask clarification questions to the user), this failure type can be significantly alleviated.

\paragraph{Handling paraphrased questions}
Questions can be semantically similar to the training samples, but lexically very dissimilar. In this case the model can have a hard time generating correct queries.
For example, given the question \emph{find the number of patients who are no longer alive}, the model must generate the condition \emph{patients.expire\_flag=1}.
This would be considerably more difficult since the model was trained with samples such as \emph{find the number of patients who expired}.
This failure type can be potentially alleviated by using a very powerful pre-trained language model.

\paragraph{Rare question types}
The model has a hard time handling question types that rarely occur in the training set.
For example, multi-part questions such as \emph{When was patient id XXX admitted? Specify time and location} occur only twice in the training set, and this provides very little chance for the model to learn to correctly answer this question type.

\vspace{-3mm}
\section{Conclusion}
\vspace{-1mm}
\label{sec:conclusion}

In this work, we proposed \model, a unified Encoder-as-Decoder model with the input masking technique to cope with EHR-QA containing complex medical terminology.
We applied \model on a large publicly available NLQ2Query dataset, MIMICSQL* and demonstrated significantly superior performance over the previous state-of-the-art method.
In addition, given the same experimental settings, our model showed consistent superior results for the graph-based EHR-QA dataset, MIMICSPARQL*.
We plan to extend our model to incorporate user interaction as discussed in the error analysis and further address more domain-specific challenges such as abbreviated terms in the future.



\acks{
We would like to thank Ping Wang and Junwoo Park for sharing codes and data resources.
We also thank anonymous reviewers for their effort and valuable feedback.
This work was supported by Institute of Information \& Communications Technology Planning \& Evaluation (IITP) grant (No.2019-0-00075, Artificial Intelligence Graduate School Program(KAIST)) and National Research Foundation of Korea (NRF) grant (NRF-2020H1D3A2A03100945), funded by the Korea government (MSIT).
}

\bibliography{jmlr-sample}

\begin{thebibliography}{29}
\providecommand{\natexlab}[1]{#1}
\providecommand{\url}[1]{\texttt{#1}}
\expandafter\ifx\csname urlstyle\endcsname\relax
  \providecommand{\doi}[1]{doi: #1}\else
  \providecommand{\doi}{doi: \begingroup \urlstyle{rm}\Url}\fi

\bibitem[Choi et~al.(2020)Choi, Shin, Kim, and Shin]{choi2020ryansql}
DongHyun Choi, Myeong~Cheol Shin, EungGyun Kim, and Dong~Ryeol Shin.
\newblock Ryansql: Recursively applying sketch-based slot fillings for complex
  text-to-sql in cross-domain databases.
\newblock \emph{arXiv preprint arXiv:2004.03125}, 2020.

\bibitem[Cucerzan and Brill(2004)]{cucerzan-brill-2004-spelling}
Silviu Cucerzan and Eric Brill.
\newblock Spelling correction as an iterative process that exploits the
  collective knowledge of web users.
\newblock In \emph{Proceedings of the 2004 Conference on Empirical Methods in
  Natural Language Processing}, pages 293--300, Barcelona, Spain, July 2004.
  Association for Computational Linguistics.
\newblock URL \url{https://aclanthology.org/W04-3238}.

\bibitem[Devlin et~al.(2019)Devlin, Chang, Lee, and Toutanova]{devlin2019bert}
Jacob Devlin, Ming-Wei Chang, Kenton Lee, and Kristina Toutanova.
\newblock Bert: Pre-training of deep bidirectional transformers for language
  understanding.
\newblock In \emph{Proceedings of the 2019 Conference of the North American
  Chapter of the Association for Computational Linguistics: Human Language
  Technologies, Volume 1 (Long and Short Papers)}, pages 4171--4186, 2019.

\bibitem[Dong and Lapata(2018)]{dong2018coarse}
Li~Dong and Mirella Lapata.
\newblock Coarse-to-fine decoding for neural semantic parsing.
\newblock In \emph{Proceedings of the 56th Annual Meeting of the Association
  for Computational Linguistics (Volume 1: Long Papers)}, pages 731--742, 2018.

\bibitem[Dong et~al.(2019)Dong, Yang, Wang, Wei, Liu, Wang, Gao, Zhou, and
  Hon]{dong2019unified}
Li~Dong, Nan Yang, Wenhui Wang, Furu Wei, Xiaodong Liu, Yu~Wang, Jianfeng Gao,
  Ming Zhou, and Hsiao-Wuen Hon.
\newblock Unified language model pre-training for natural language
  understanding and generation.
\newblock In \emph{Proceedings of the 33rd International Conference on Neural
  Information Processing Systems}, pages 13063--13075, 2019.

\bibitem[Giordani and Moschitti(2012)]{giordani-moschitti-2012-translating}
Alessandra Giordani and Alessandro Moschitti.
\newblock Translating questions to {SQL} queries with generative parsers
  discriminatively reranked.
\newblock In \emph{Proceedings of {COLING} 2012: Posters}, pages 401--410,
  Mumbai, India, December 2012. The COLING 2012 Organizing Committee.
\newblock URL \url{https://www.aclweb.org/anthology/C12-2040}.

\bibitem[Guo et~al.(2019)Guo, Zhan, Gao, Xiao, Lou, Liu, and
  Zhang]{guo2019towards}
Jiaqi Guo, Zecheng Zhan, Yan Gao, Yan Xiao, Jian-Guang Lou, Ting Liu, and
  Dongmei Zhang.
\newblock Towards complex text-to-sql in cross-domain database with
  intermediate representation.
\newblock In \emph{Proceedings of the 57th Annual Meeting of the Association
  for Computational Linguistics}, pages 4524--4535, 2019.

\bibitem[Hagiwara and Mita(2020)]{hagiwara-mita-2020-github}
Masato Hagiwara and Masato Mita.
\newblock {G}it{H}ub typo corpus: A large-scale multilingual dataset of
  misspellings and grammatical errors.
\newblock In \emph{Proceedings of the 12th Language Resources and Evaluation
  Conference}, pages 6761--6768, Marseille, France, May 2020. European Language
  Resources Association.
\newblock ISBN 979-10-95546-34-4.
\newblock URL \url{https://aclanthology.org/2020.lrec-1.835}.

\bibitem[Hwang et~al.(2019)Hwang, Yim, Park, and Seo]{hwang2019comprehensive}
Wonseok Hwang, Jinyeong Yim, Seunghyun Park, and Minjoon Seo.
\newblock A comprehensive exploration on wikisql with table-aware word
  contextualization.
\newblock \emph{arXiv preprint arXiv:1902.01069}, 2019.

\bibitem[Johnson et~al.(2016)Johnson, Pollard, Shen, Li-Wei, Feng, Ghassemi,
  Moody, Szolovits, Celi, and Mark]{johnson2016mimic}
Alistair~EW Johnson, Tom~J Pollard, Lu~Shen, H~Lehman Li-Wei, Mengling Feng,
  Mohammad Ghassemi, Benjamin Moody, Peter Szolovits, Leo~Anthony Celi, and
  Roger~G Mark.
\newblock Mimic-iii, a freely accessible critical care database.
\newblock \emph{Scientific data}, 3\penalty0 (1):\penalty0 1--9, 2016.

\bibitem[Kemighan et~al.(1990)Kemighan, Church, and Gale]{kemighan1990spelling}
Mark~D Kemighan, Kenneth Church, and William~A Gale.
\newblock A spelling correction program based on a noisy channel model.
\newblock In \emph{COLING 1990 Volume 2: Papers presented to the 13th
  International Conference on Computational Linguistics}, 1990.

\bibitem[Kingma and Ba(2014)]{kingma2014adam}
Diederik~P Kingma and Jimmy Ba.
\newblock Adam: A method for stochastic optimization.
\newblock \emph{arXiv preprint arXiv:1412.6980}, 2014.

\bibitem[Li and Jagadish(2014)]{li2014constructing}
Fei Li and HV~Jagadish.
\newblock Constructing an interactive natural language interface for relational
  databases.
\newblock \emph{Proceedings of the VLDB Endowment}, 8\penalty0 (1):\penalty0
  73--84, 2014.

\bibitem[Luong et~al.(2015)Luong, Pham, and Manning]{luong2015effective}
Minh-Thang Luong, Hieu Pham, and Christopher~D Manning.
\newblock Effective approaches to attention-based neural machine translation.
\newblock In \emph{Proceedings of the 2015 Conference on Empirical Methods in
  Natural Language Processing}, pages 1412--1421, 2015.

\bibitem[Pampari et~al.(2018)Pampari, Raghavan, Liang, and
  Peng]{pampari2018emrqa}
Anusri Pampari, Preethi Raghavan, Jennifer Liang, and Jian Peng.
\newblock emrqa: A large corpus for question answering on electronic medical
  records.
\newblock In \emph{Proceedings of the 2018 Conference on Empirical Methods in
  Natural Language Processing}, pages 2357--2368, 2018.

\bibitem[Park et~al.(2021)Park, Cho, Lee, Choo, and Choi]{park2021knowledge}
Junwoo Park, Youngwoo Cho, Haneol Lee, Jaegul Choo, and Edward Choi.
\newblock Knowledge graph-based question answering with electronic health
  records.
\newblock In \emph{Machine Learning for Healthcare Conference}, pages 36--53.
  PMLR, 2021.

\bibitem[Pollard et~al.(2018)Pollard, Johnson, Raffa, Celi, Mark, and
  Badawi]{pollard2018eicu}
Tom~J Pollard, Alistair~EW Johnson, Jesse~D Raffa, Leo~A Celi, Roger~G Mark,
  and Omar Badawi.
\newblock The eicu collaborative research database, a freely available
  multi-center database for critical care research.
\newblock \emph{Scientific data}, 5\penalty0 (1):\penalty0 1--13, 2018.

\bibitem[Price(1990)]{price1990evaluation}
Patti Price.
\newblock Evaluation of spoken language systems: The atis domain.
\newblock In \emph{Speech and Natural Language: Proceedings of a Workshop Held
  at Hidden Valley, Pennsylvania, June 24-27, 1990}, 1990.

\bibitem[Quirk et~al.(2015)Quirk, Mooney, and Galley]{quirk2015language}
Chris Quirk, Raymond Mooney, and Michel Galley.
\newblock Language to code: Learning semantic parsers for if-this-then-that
  recipes.
\newblock In \emph{Proceedings of the 53rd Annual Meeting of the Association
  for Computational Linguistics and the 7th International Joint Conference on
  Natural Language Processing (Volume 1: Long Papers)}, pages 878--888, 2015.

\bibitem[Raghavan et~al.(2021)Raghavan, Liang, Mahajan, Chandra, and
  Szolovits]{raghavan2021emrkbqa}
Preethi Raghavan, Jennifer~J Liang, Diwakar Mahajan, Rachita Chandra, and Peter
  Szolovits.
\newblock emrkbqa: A clinical knowledge-base question answering dataset.
\newblock In \emph{Proceedings of the 20th Workshop on Biomedical Language
  Processing}, pages 64--73, 2021.

\bibitem[Rothe et~al.(2020)Rothe, Narayan, and Severyn]{rothe2020leveraging}
Sascha Rothe, Shashi Narayan, and Aliaksei Severyn.
\newblock Leveraging pre-trained checkpoints for sequence generation tasks.
\newblock \emph{Transactions of the Association for Computational Linguistics},
  8:\penalty0 264--280, 2020.

\bibitem[{\v{S}}uster and Daelemans(2018)]{vsuster2018clicr}
Simon {\v{S}}uster and Walter Daelemans.
\newblock Clicr: A dataset of clinical case reports for machine reading
  comprehension.
\newblock In \emph{Proceedings of NAACL-HLT}, pages 1551--1563, 2018.

\bibitem[Vaswani et~al.(2017)Vaswani, Shazeer, Parmar, Uszkoreit, Jones, Gomez,
  Kaiser, and Polosukhin]{vaswani2017attention}
Ashish Vaswani, Noam Shazeer, Niki Parmar, Jakob Uszkoreit, Llion Jones,
  Aidan~N Gomez, Lukasz Kaiser, and Illia Polosukhin.
\newblock Attention is all you need.
\newblock \emph{arXiv preprint arXiv:1706.03762}, 2017.

\bibitem[Wang et~al.(2020{\natexlab{a}})Wang, Shin, Liu, Polozov, and
  Richardson]{wang2020rat}
Bailin Wang, Richard Shin, Xiaodong Liu, Oleksandr Polozov, and Matthew
  Richardson.
\newblock Rat-sql: Relation-aware schema encoding and linking for text-to-sql
  parsers.
\newblock In \emph{Proceedings of the 58th Annual Meeting of the Association
  for Computational Linguistics}, pages 7567--7578, 2020{\natexlab{a}}.

\bibitem[Wang et~al.(2020{\natexlab{b}})Wang, Shi, and Reddy]{wang2020text}
Ping Wang, Tian Shi, and Chandan~K Reddy.
\newblock Text-to-sql generation for question answering on electronic medical
  records.
\newblock In \emph{Proceedings of The Web Conference 2020}, pages 350--361,
  2020{\natexlab{b}}.

\bibitem[Wu et~al.(2016)Wu, Schuster, Chen, Le, Norouzi, Macherey, Krikun, Cao,
  Gao, Macherey, et~al.]{wu2016google}
Yonghui Wu, Mike Schuster, Zhifeng Chen, Quoc~V Le, Mohammad Norouzi, Wolfgang
  Macherey, Maxim Krikun, Yuan Cao, Qin Gao, Klaus Macherey, et~al.
\newblock Google's neural machine translation system: Bridging the gap between
  human and machine translation.
\newblock \emph{arXiv preprint arXiv:1609.08144}, 2016.

\bibitem[Yu et~al.(2018)Yu, Zhang, Yang, Yasunaga, Wang, Li, Ma, Li, Yao,
  Roman, et~al.]{yu2018spider}
Tao Yu, Rui Zhang, Kai Yang, Michihiro Yasunaga, Dongxu Wang, Zifan Li, James
  Ma, Irene Li, Qingning Yao, Shanelle Roman, et~al.
\newblock Spider: A large-scale human-labeled dataset for complex and
  cross-domain semantic parsing and text-to-sql task.
\newblock In \emph{Proceedings of the 2018 Conference on Empirical Methods in
  Natural Language Processing}, pages 3911--3921, 2018.

\bibitem[Zhong et~al.(2017)Zhong, Xiong, and Socher]{zhong2017seq2sql}
Victor Zhong, Caiming Xiong, and Richard Socher.
\newblock Seq2sql: Generating structured queries from natural language using
  reinforcement learning.
\newblock \emph{arXiv preprint arXiv:1709.00103}, 2017.

\bibitem[Zhu et~al.(2020)Zhu, Ahuja, Juan, Wei, and Reddy]{zhu2020question}
Ming Zhu, Aman Ahuja, Da-Cheng Juan, Wei Wei, and Chandan~K Reddy.
\newblock Question answering with long multiple-span answers.
\newblock In \emph{Proceedings of the 2020 Conference on Empirical Methods in
  Natural Language Processing: Findings}, pages 3840--3849, 2020.

\end{thebibliography}

\clearpage
\appendix
\section{Implementation Details}\label{apd:first}
\label{appendix:implement}

\subsection{Noise generator}
\vspace{-3mm}
\begin{algorithm2e}[h]
\caption{Noise Generator}
\label{alg:noise_gen}
\KwIn{a question $Q = \{ q_1,q_2,\ldots,q_n\}$, noise rate $r_{\textsf{noise}}$, min word length $l_{\textsf{min}}$ } 
\KwOut{a noisy question $\tilde{Q} = \{ \tilde{q_1},\tilde{q_2},\ldots,\tilde{q_n} \}$ }
$\tilde{Q}$ $\leftarrow$ [ ]\\
\For{$i \leftarrow 1$ \KwTo $n$}{
    $p \leftarrow \text{Uniform}(0, 1)$ \\
    \uIf{
            $p \cdot \log \textsc{[} \textsc{len(}q_i) \textsc{]} \leq r_{\textsf{noise}}$
        }
        {
        $r \leftarrow \text{Uniform}(0, 1)$ \\
        \uIf{
            \textsc{(IsNumeric($q_i$)} \textbf{or} \textsc{IsDate($q_i$)} \textbf{or} \textsc{len}($q_i$) $\leq$ $l_{\textsf{min}}$)
        }{
        $\tilde{q_i}$ $\leftarrow$ $q_i$
        }
        \uElseIf{$0$ $\leq$ r $<$ $0.15$}{
        $\tilde{q_i}$  $\leftarrow$ \textsc{ExtraLetter($q_i$)}
        }
        \uElseIf{$0.15$ $\leq$ r $<$ $0.30$}{
        $\tilde{q_i}$  $\leftarrow$ \textsc{MissingLetter($q_i$)}
        }
        \uElseIf{$0.30$ $\leq$ r $<$ $0.50$}{
        $\tilde{q_i}$  $\leftarrow$ \textsc{WrongLetter($q_i$)}
        }
        \Else{
        $\tilde{q_i}$  $\leftarrow$ \textsc{ReversedLetter($q_i$)}
        }
    }
    \Else{
    $\tilde{q_i}$  $\leftarrow$ $q_i$
    }
    $\tilde{Q}$.\textsc{Add($\tilde{q_i}$)}
}\end{algorithm2e}
\vspace{-5mm}

\subsection{Implementation details}
\vspace{-1mm}
We implement our model and baseline models with PyTorch Lightning \footnote{\url{https://www.pytorchlightning.ai}} and HuggingFace's transformers\footnote{\url{https://huggingface.co/transformers/}}. In the case of TREQS, we utilized the official code\footnote{\url{https://github.com/wangpinggl/TREQS}} written by the origin authors.
We use the original BERT as our pre-trained model.
For the fair comparison, we adjust the number of self-attention layer of encoder and decoder in Encoder-to-Decoder model to half of ours.
We use Adam optimizer \citep{kingma2014adam} with the learning rate set to $3\times10^{\text{-}5}$ and batch size set to 32. 

\subsection{Hyperparameters}
\vspace{-1mm}
In order to make an accurate comparison with the baseline models, Seq2seq model and TREQS model were imported from \cite{park2021knowledge}, and hyperparameters were also imported with the same value. We trained our models on two types of GPU environments: NVIDIA Tesla T4 and NVIDIA GeForce RTX-3090. Also, torch version is 1.7.0, and CUDA version is 11.1. Other hyparparameters are presented in Table \ref{tab:hparams}.

\begin{table*}[ht]
\resizebox{\textwidth}{!}{%
\begin{tabular}{ccccc}
\hline
Hyperparameters & Seq2Seq & TREQS & E-to-D (+IM) & E-as-D (+IM) \\ \hline
Hidden dimension & 256 & 256(enc) + 256(dec) & 768 & 768\\
Learning rate & \num{5e-4}  & \num{5e-4} & \num{3e-5} & \num{3e-5} \\
LR Scheduler & \begin{tabular}[c]{@{}c@{}}StepLR(step size = 2,\\ step decay = 0.8)\end{tabular} & \begin{tabular}[c]{@{}c@{}}StepLR(step size = 2,\\ step decay = 0.8)\end{tabular} & Linear decay & Linear decay \\
Batch size & 16 & 64 & 32 & 32\\
Epochs & 20 & 20 & 100 (w/ early stop) & 100 (w/ early stop) \\
Seed & 1, 12, 123, 1234, 42 & 1, 12, 123, 1234, 42 & 1, 12, 123, 1234, 42  & 1, 12, 123, 1234, 42  \\
Beam size  & - & 5 & 5 & 5 \\
Seq2seq-LM prob.& - & -& - & 0.3\\
Input Masking prob. & - & - & 0.2 (only for IM) & 0.2 (only for IM)\\ \hline
\end{tabular}
}
\vspace{-3mm}
\caption{Hyperparameters for training several models.}
\label{tab:hparams}
\end{table*}

\section{Results}\label{apd:second}


\subsection{Quantitative results of noisy MIMICSPARQL*}
We provide the quantitative results on MIMICSPARQL* dataset with the different degree of noise in the Table~\ref{tab:noise_sparql_exp}.
\begin{table*}[t]
\centering
\resizebox{\textwidth}{!}{
\begin{tabular}{llllllllll}
\hline
{\textbf{Method}} &
  \multicolumn{3}{c}{\textbf{Testing-weak (5\% typo prob.)}} &
  \multicolumn{3}{c}{\textbf{Testing-moderate (10\% typo prob.)}} &
  \multicolumn{3}{c}{\textbf{Testing-strong (15\% typo prob.)}} \\
 &
  \multicolumn{1}{c}{$Acc_{LF}$} &
  \multicolumn{1}{c}{$Acc_{EX}$} &
  \multicolumn{1}{c}{$Acc_{ST}$} &
  \multicolumn{1}{c}{$Acc_{LF}$} &
  \multicolumn{1}{c}{$Acc_{EX}$} &
  \multicolumn{1}{c}{$Acc_{ST}$} &
  \multicolumn{1}{c}{$Acc_{LF}$} &
  \multicolumn{1}{c}{$Acc_{EX}$} &
  \multicolumn{1}{c}{$Acc_{ST}$} \\ \hline
\textbf{Before Recovering} \\
Seq2Seq &
  $0.159\ {\scriptstyle (0.039)}$ &
  $0.265\ {\scriptstyle (0.029)}$ &
  $0.260\ {\scriptstyle (0.011)}$ &
  $0.093\ {\scriptstyle (0.016)}$ &
  $0.178\ {\scriptstyle (0.019)}$ &
  $0.178\ {\scriptstyle (0.019)}$ &
  $0.070\ {\scriptstyle (0.016)}$ &
  $0.140\ {\scriptstyle (0.018)}$ &
  $0.127\ {\scriptstyle (0.014)}$ \\
TREQS &
  $0.419\ {\scriptstyle (0.020)}$ &
  $0.551\ {\scriptstyle (0.015)}$ &
  $0.630\ {\scriptstyle (0.018)}$ &
  $0.277\ {\scriptstyle (0.016)}$ &
  $0.426\ {\scriptstyle (0.009)}$ &
  $0.494\ {\scriptstyle (0.019)}$ &
  $0.215\ {\scriptstyle (0.009)}$ &
  $0.376\ {\scriptstyle (0.011)}$ &
  $0.422\ {\scriptstyle (0.015)}$ \\
E-to-D &
  $0.704\ {\scriptstyle (0.011)}$ &
  $0.778\ {\scriptstyle (0.010)}$ &
  $0.833\ {\scriptstyle (0.013)}$ &
  $0.547\ {\scriptstyle (0.009)}$ &
  $0.639\ {\scriptstyle (0.010)}$ &
  $0.704\ {\scriptstyle (0.016)}$ &
  $0.449\ {\scriptstyle (0.020)}$ &
  $0.549\ {\scriptstyle (0.019)}$ &
  $0.643\ {\scriptstyle (0.029)}$ \\
E-to-D + IM &
  $0.679\ {\scriptstyle (0.015)}$ &
  $0.764\ {\scriptstyle (0.019)}$ &
  $0.801\ {\scriptstyle (0.015)}$ &
  $0.534\ {\scriptstyle (0.025)}$ &
  $0.631\ {\scriptstyle (0.031)}$ &
  $0.678\ {\scriptstyle (0.032)}$ &
  $0.465\ {\scriptstyle (0.012)}$ &
  $0.565\ {\scriptstyle (0.009)}$ &
  $0.621\ {\scriptstyle (0.020)}$ \\
E-as-D &
  $0.660\ {\scriptstyle (0.006)}$ &
  $0.741\ {\scriptstyle (0.005)}$ &
  $0.837\ {\scriptstyle (0.014)}$ &
  $0.504\ {\scriptstyle (0.006)}$ &
  $0.609\ {\scriptstyle (0.010)}$ &
  $0.715\ {\scriptstyle (0.024)}$ &
  $0.385\ {\scriptstyle (0.008)}$ &
  $0.493\ {\scriptstyle (0.014)}$ &
  $0.648\ {\scriptstyle (0.030)}$ \\
UniQA (E-as-D + IM) &
  $\textbf{0.754}\ {\scriptstyle (0.022)}$ &
  $\textbf{0.820}\ {\scriptstyle (0.018)}$ &
  $\textbf{0.877}\ {\scriptstyle (0.013)}$ &
  $\textbf{0.646}\ {\scriptstyle (0.015)}$ &
  $\textbf{0.725}\ {\scriptstyle (0.015)}$ &
  $\textbf{0.798}\ {\scriptstyle (0.016)}$ &
  $\textbf{0.535}\ {\scriptstyle (0.022)}$ &
  $\textbf{0.627}\ {\scriptstyle (0.023)}$ &
  $\textbf{0.726}\ {\scriptstyle (0.014)}$ \\ \hline
\textbf{After Recovering} \\
Seq2Seq &
  $0.160\ {\scriptstyle (0.040)}$ &
  $0.273\ {\scriptstyle (0.030)}$ &
  $0.260\ {\scriptstyle (0.011)}$ &
  $0.094\ {\scriptstyle (0.017)}$ &
  $0.182\ {\scriptstyle (0.020)}$ &
  $0.178\ {\scriptstyle (0.019)}$ &
  $0.071\ {\scriptstyle (0.016)}$ &
  $0.143\ {\scriptstyle (0.017)}$ &
  $0.127\ {\scriptstyle (0.014)}$ \\
TREQS &
  $0.426\ {\scriptstyle (0.021)}$ &
  $0.562\ {\scriptstyle (0.017)}$ &
  $0.630\ {\scriptstyle (0.018)}$ &
  $0.283\ {\scriptstyle (0.017)}$ &
  $0.434\ {\scriptstyle (0.008)}$ &
  $0.494\ {\scriptstyle (0.019)}$ &
  $0.221\ {\scriptstyle (0.009)}$ &
  $0.382\ {\scriptstyle (0.009)}$ &
  $0.422\ {\scriptstyle (0.015)}$ \\
E-to-D &
  $0.777\ {\scriptstyle (0.015)}$ &
  $0.847\ {\scriptstyle (0.013)}$ &
  $0.833\ {\scriptstyle (0.013)}$ &
  $0.618\ {\scriptstyle (0.017)}$ &
  $0.708\ {\scriptstyle (0.021)}$ &
  $0.704\ {\scriptstyle (0.016)}$ &
  $0.526\ {\scriptstyle (0.028)}$ &
  $0.620\ {\scriptstyle (0.028)}$ &
  $0.643\ {\scriptstyle (0.029)}$ \\
E-to-D + IM &
  $0.696\ {\scriptstyle (0.012)}$ &
  $0.786\ {\scriptstyle (0.016)}$ &
  $0.801\ {\scriptstyle (0.015)}$ &
  $0.546\ {\scriptstyle (0.024)}$ &
  $0.644\ {\scriptstyle (0.029)}$ &
  $0.678\ {\scriptstyle (0.032)}$ &
  $0.481\ {\scriptstyle (0.011)}$ &
  $0.582\ {\scriptstyle (0.010)}$ &
  $0.621\ {\scriptstyle (0.020)}$ \\
E-as-D &
  $0.819\ {\scriptstyle (0.012)}$ &
  $0.874\ {\scriptstyle (0.016)}$ &
  $0.837\ {\scriptstyle (0.014)}$ &
  $0.668\ {\scriptstyle (0.018)}$ &
  $0.740\ {\scriptstyle (0.022)}$ &
  $0.715\ {\scriptstyle (0.024)}$ &
  $0.571\ {\scriptstyle (0.022)}$ &
  $0.637\ {\scriptstyle (0.031)}$ &
  $0.648\ {\scriptstyle (0.030)}$ \\
UniQA (E-as-D + IM) &
  $\textbf{0.835}\ {\scriptstyle (0.012)}$ &
  $\textbf{0.896}\ {\scriptstyle (0.012)}$ &
  $\textbf{0.877}\ {\scriptstyle (0.013)}$ &
  $\textbf{0.725}\ {\scriptstyle (0.010)}$ &
  $\textbf{0.799}\ {\scriptstyle (0.011)}$ &
  $\textbf{0.798}\ {\scriptstyle (0.016)}$ &
  $\textbf{0.625}\ {\scriptstyle (0.009)}$ &
  $\textbf{0.706}\ {\scriptstyle (0.007)}$ &
  $\textbf{0.726}\ {\scriptstyle (0.014)}$ \\ \hline
\end{tabular}%
}
\vspace{-1mm}
\caption{QA Performance on NOISY MIMICSPARQL* natural questions with evaluated with logic form accuracy ($Acc_{LF}$), execution accuracy ($Acc_{EX}$), and the structural accuracy ($Acc_{ST}$). Based on the probability of noise, we refer to 5\% as weak, 10\% as moderate, and 15\% as strong.}
\label{tab:noise_sparql_exp}
\end{table*}

\subsection{Qualitative results of MIMICSQL*}
We provide the qualitative results on MIMICSQL* dataset in the Figure~\ref{fig:base_qual}.
\begin{figure*}[ht]
\centering
\resizebox{\textwidth}{!}{
    \begin{tabular}{p{2.5cm}|p{23cm}}
    \toprule
    NLQ
    & \text{how many patients with  elective admission type had the procedure titled \textbf{other operations on heart and pericardium}?}
    \\ \midrule
    Ground \newline Truth 
    & \text{select count (distinct patients.subject\_id) from patients inner join admissions on patients.subject\_id = admissions.subject\_id inner join} \newline 
    \text{procedures on admissions.hadm\_id = procedures.hadm\_id inner join d\_icd\_procedures on procedures.icd9\_code = d\_icd\_procedures.icd9\_code} \newline
    \text{where admissions.admission\_type = ``elective'' and d\_icd\_procedures.long\_title = \textbf{``other operations on heart and pericardium''}} \\
    \midrule
    Seq2Seq
    & \text{select count (distinct patients.subject\_id) from patients inner join admissions on patients.subject\_id = admissions.subject\_id inner join} \newline
    \text{procedures on admissions.hadm\_id = procedures.hadm\_id inner join d\_icd\_procedures on procedures.icd9\_code = d\_icd\_procedures.icd9\_code} \newline
    \text{where admissions.admission\_type = ``elective'' and d\_icd\_procedures.short\_title = \textbf{``other} \boldred{$<$UNK$>$ nec}\textbf{''}} \\ 
    \hline
    TREQS
    & \text{select count (distinct patients.subject\_id) from patients inner join admissions on patients.subject\_id = admissions.subject\_id inner join} \newline
    \text{procedures on admissions.hadm\_id = procedures.hadm\_id inner join d\_icd\_procedures on procedures.icd9\_code = d\_icd\_procedures.icd9\_code} \newline
    \text{where admissions.admission\_type = ``elective'' and d\_icd\_procedures.long\_title = \textbf{``other operations on} \boldred{on} \textbf{heart and pericardium''}} \\
    \hline
    E-to-D 
    & \text{select count (distinct patients.subject\_id) from patients inner join admissions on patients.subject\_id = admissions.subject\_id inner join} \newline
    \text{procedures on admissions.hadm\_id = procedures.hadm\_id inner join d\_icd\_procedures on procedures.icd9\_code = d\_icd\_procedures.icd9\_code} \newline
    \text{where admissions.admission\_type = ``elective'' and d\_icd\_procedures.long\_title = \textbf{``other} \boldred{conversion of} \textbf{heart and pericardium''}} \\
    \hline
    E-to-D + IM 
    & \text{select count (distinct patients.subject\_id) from patients inner join admissions on patients.subject\_id = admissions.subject\_id inner join} \newline
    \text{procedures on admissions.hadm\_id = procedures.hadm\_id inner join d\_icd\_procedures on procedures.icd9\_code = d\_icd\_procedures.icd9\_code} \newline
    \text{where admissions.admission\_type = ``elective'' and d\_icd\_procedures.long\_title = \textbf{``other} \boldred{diagnostic procedures on skin and subcutaneous tissue}\textbf{''}} \\
    \hline
    E-as-D 
    & \text{select count (distinct patients.subject\_id) from patients inner join admissions on patients.subject\_id = admissions.subject\_id inner join} \newline
    \text{procedures on admissions.hadm\_id = procedures.hadm\_id inner join d\_icd\_procedures on procedures.icd9\_code = d\_icd\_procedures.icd9\_code} \newline
    \text{where admissions.admission\_type = ``elective'' and d\_icd\_procedures.long\_title = \textbf{``other operations on heart and pericardium''}} \\
    \hline
    UniQA \newline(E-as-D + IM) 
    & \text{select count (distinct patients.subject\_id) from patients inner join admissions on patients.subject\_id = admissions.subject\_id inner join} \newline
    \text{procedures on admissions.hadm\_id = procedures.hadm\_id inner join d\_icd\_procedures on procedures.icd9\_code = d\_icd\_procedures.icd9\_code} \newline
    \text{where admissions.admission\_type = ``elective'' and d\_icd\_procedures.long\_title = \textbf{``other operations on heart and pericardium''}} \\
    \bottomrule
    \end{tabular}
}
\vspace{-1mm}
\caption{SQL Queries generated by different models given the same NLQ in MIMICSQL* test set. 
Conditional values corresponding to the medical term ``\textit{other operations on heart and pericardium}'' mentioned in the NLQ are marked in bold. The incorrectly predicted tokens are highlighted in red color.}
\label{fig:base_qual}
\end{figure*}

\end{document}